%% file: DeepInterestNetwork.tex
\documentclass[sigconf]{acmart}
\usepackage{amsmath,amssymb,amsfonts,bm,amsthm}
\usepackage[linesnumbered,ruled]{algorithm2e}
\usepackage{graphicx}
\usepackage[tight,normal]{subfigure}
\usepackage{multirow,makecell,url,footmisc}
\usepackage{enumerate}
\usepackage{lineno}
\usepackage{textcomp}
\usepackage{booktabs} % For formal tables
\usepackage{subfigure}
\usepackage{amssymb}
\usepackage{amsmath}
\usepackage{booktabs}
\usepackage{bm}
\usepackage{mathtools}
\usepackage{threeparttable}
\usepackage[american]{babel}
\usepackage{color}
\usepackage{threeparttable}
\usepackage{appendix}
\usepackage{comment}
\usepackage{multirow}
\usepackage{blindtext}
\usepackage{comment}

\newcommand{\bs}{\boldsymbol}

\usepackage{comment}

\renewcommand\footnotetextcopyrightpermission[1]{}
\copyrightyear{2018} 
\acmYear{2018} 
\setcopyright{acmcopyright}
\acmConference{KDD'18}{}{August 19--23, 2018, London, United Kingdom.}

\begin{document}

\title{Deep Interest Network for Click-Through Rate Prediction}

\author{Guorui Zhou, Chengru Song, Xiaoqiang Zhu \\
  Ying Fan, Han Zhu, Xiao Ma, Yanghui Yan, Junqi Jin, Han Li, Kun Gai}
\affiliation{%
  \institution{Alibaba Group}
}
%\affiliation{$^\dag$Alibaba Inc.}
\email{{guorui.xgr, chengru.scr, xiaoqiang.zxq, zhuhan.zh, fanying.fy, maxiao.ma, yanghui.yyh, junqi.jjq, lihan.hl,jingshi.gk}@alibaba-inc.com}

\begin{abstract}

Click-through rate prediction is an essential task in industrial applications, such as online advertising.
Recently deep learning based models have been proposed, which follow a similar Embedding\&MLP paradigm.
In these methods large scale sparse input features are first mapped into low dimensional embedding vectors, and then transformed into fixed-length vectors in a group-wise manner, finally concatenated together to fed into a multilayer perceptron (MLP) to learn the nonlinear relations among features.
In this way, user features are compressed into a fixed-length representation vector, in regardless of what candidate ads are.
The use of fixed-length vector will be a bottleneck, which brings difficulty  for Embedding\&MLP methods to capture user's diverse interests effectively from rich historical behaviors.
%The use of fixed-length vector will be a bottleneck to express user's diverse interests. This brings difficulty   for Embedding\&MLP methods to capture user's diverse interests effectively from rich historical behaviors.
%The use of fixed-length vector will be a bottleneck to express user's diverse interests, limiting the effectiveness of Embedding\&MLP methods to capture user interests from rich historical behaviors. 
In this paper, we propose a novel model: Deep Interest Network (DIN) which tackles this challenge by designing a local activation unit to adaptively learn the representation of user interests from historical behaviors with respect to a certain ad.
This representation vector varies over different ads, improving the expressive ability of model greatly.
Besides, we develop two techniques: mini-batch aware regularization and data adaptive activation function which can help training industrial deep networks with hundreds of millions of parameters.
Experiments on two public datasets as well as an Alibaba real production dataset with over 2 billion samples demonstrate the effectiveness of proposed approaches, which achieve superior performance compared with state-of-the-art methods.
%DIN now has been successfully deployed in the productive displaying system in Alibaba, contributing up to 10.0\% CTR promotion which is a significant improvement to the business.
DIN now has been successfully deployed in the online display advertising system in Alibaba, serving the main traffic.

\end{abstract}
\begin{CCSXML}
<ccs2012>
<concept>
<concept_id>10002951.10003260.10003272.10003275</concept_id>
<concept_desc>Information systems~Display advertising</concept_desc>
<concept_significance>500</concept_significance>
</concept>
<concept>
<concept_id>10002951.10003317.10003347.10003350</concept_id>
<concept_desc>Information systems~Recommender systems</concept_desc>
<concept_significance>500</concept_significance>
</concept>
</ccs2012>
\end{CCSXML}

\ccsdesc[500]{Information systems~Display advertising}
\ccsdesc[500]{Information systems~Recommender systems}

\keywords{Click-Through Rate Prediction, Display Advertising, E-commerce}
%\keywords{Click-Through Rate Prediction, Display Advertising, E-commerce, User's Diverse Interests, Training Industrial Deep Network}
\maketitle

\input{ch_intro}
\input{ch_relatedwork}
\input{ch_sysview}

\input{ch_approach}
\input{ch_exp}

\input{ch_concln}
\bibliographystyle{ACM-Reference-Format}
\bibliography{DIN} 
\end{document}

%% file: ch_intro.tex
\section{Introduction}
In cost-per-click (CPC) advertising system, advertisements are ranked by the eCPM (effective cost per mille), which is the product of the bid price and CTR (click-through rate), and CTR needs to be predicted by the system. Hence, the performance of CTR prediction model has a direct impact on the final revenue and plays a key role in the advertising system. Modeling CTR prediction has received much attention from both research and industry community.

Recently, inspired by the success of deep learning in computer vision~\cite{huang2016densely} and natural language processing~\cite{bengio:attention}, deep learning based methods have been proposed for CTR prediction task \cite{deep_crossing,youtube:recommend,deep_intent,widedeep}.
These methods follow a similar Embedding\&MLP paradigm: large scale sparse input features are first mapped into low dimensional embedding vectors, and then transformed into fixed-length vectors in a group-wise manner, finally concatenated together to fed into fully connected layers (also known as multilayer perceptron, MLP) to learn the nonlinear relations among features. 
Compared with commonly used logistic regression model \cite{ftrl}, these deep learning methods can reduce a lot of feature engineering jobs and enhance the model capability greatly. For simplicity, we name these methods Embedding\&MLP in this paper, which now have become popular on CTR prediction task.

However, the user representation vector with a limited dimension in Embedding\&MLP methods will be a bottleneck to express user's diverse interests. %, limiting the effectiveness of model.
Take display advertising in e-commerce site as an example. Users might be interested in different kinds of goods simultaneously when visiting the e-commerce site. That is to say, user interests are \textsl{\textbf{diverse}}. When it comes to CTR prediction task, user interests are usually captured from user behavior data. 
Embedding\&MLP methods learn the  representation of all interests for a certain user by transforming the embedding vectors of user behaviors into a fixed-length vector, which is in an euclidean space where all users' representation vectors are. 
In other words, diverse interests of the user are compressed into a fixed-length vector, which limits the expressive ability of Embedding\&MLP methods. 
%Embedding\&MLP methods learn the  representation of all interests for a certain user by transforming the embedding vectors of user behaviors into a fixed-length vector.
%In other words, diverse interests of the user are compressed into this fixed-length vector, which limits the expressive ability of Embedding\&MLP methods. 
%It is difficult to compress all the interests of a certain user into a fixed-length vector. 
%It is challengeable to compress all the interests of a user into a fixed-length vector.  
%The use of fixed-length vector to compress all the interests  
%gives the rein to the expressing, and Embedding\&MLP methods may not rise to the challenge: capturing diverse interests of user from her behaviors. 
%It is difficult to compress all the interests of user into a fixed-length vector. 
To make the representation capable enough for expressing user's diverse interests,  the dimension of the fixed-length vector needs to be largely expanded. 
Unfortunately, it will dramatically enlarge the size of learning parameters and aggravate the risk of overfitting under limited data. Besides, it adds the burden of computation and storage, which may not be tolerated for an industrial online system. 

On the other hand, it is not necessary to compress all the diverse interests of a certain user into the same vector when predicting a candidate ad because only part of user's interests will influence his/her action (to click or not to click). For example, a female swimmer will click a recommended goggle mostly due to the bought of bathing suit rather than the shoes in her last week's shopping list. 
Motivated by this, we propose a novel model: Deep Interest Network (DIN), which adaptively calculates the representation vector of user interests by taking into consideration the relevance of historical behaviors given a candidate ad. 
By introducing a local activation unit, DIN pays attentions to the related user interests by soft-searching for relevant parts of historical behaviors and takes a weighted sum pooling to obtain the representation of user interests with respect to the candidate ad.  
Behaviors with higher relevance to the candidate ad get higher activated weights and dominate the representation of user interests. 
We visualize this phenomenon in the experiment section. 
In this way, the representation vector of user interests varies over different ads, which improves the expressive ability of model under limited dimension and enables DIN to better capture user's diverse interests.

Training industrial deep networks with large scale sparse features is of great challenge. 
For example, SGD based optimization methods only update those parameters of sparse features appearing in each mini-batch. However, adding with traditional $\ell_2$ regularization, the computation turns to be unacceptable, which needs to calculate L2-norm over the whole parameters (with size scaling up to billions in our situation) for each mini-batch. In this paper, we develop a novel mini-batch aware regularization where only parameters of non-zero features appearing in each mini-batch participate in the calculation of L2-norm, making the computation acceptable. Besides, we design a data adaptive activation function, which generalizes commonly used PReLU\cite{ms:prelu} by adaptively adjusting the rectified point w.r.t. distribution of inputs and is shown to be helpful for training industrial networks with sparse features.

%By exploiting the the characteristic of sparse 
%In this paper, we carefully study the characteristic of training deep networks with sparse inputs and develop two practical techniques: mini-batch aware regularization and data adaptive activation function, which is useful to help improving the performance of trained model. 
%For example, size of parameters of network trained on our productive dataset scales up to hundreds of millions. 
%It is not an easy job to converge to a good and stable solution when training such deep networks.   
%In this paper, we introduce two practical techniques: mini-batch aware regularization and data adaptive activation function, which is useful to help improving the performance of trained model.     

%Experiments on two public datasets as well as a dataset collected from the online display advertising system in Alibaba demonstrate the effectiveness of proposed approaches, which achieve superior performance compared with state-of-the-art methods on CTR prediction task. DIN trained with the proposed regularization and activation function has been deployed in the display advertising system in Alibaba, contributing up to 10.0\% CTR promotion, which is a significant improvement to the business.

The contributions of this paper are summarized as follows:
%\begin{itemize}
%\item We point out the diversity of user interests and the use of fixed-length vector will be a bottleneck to express user's diverse interests, as well as propose deep interest network (DIN) to better capture this characteristic. We design interest activation unit in DIN to adaptively learn the representation of user interests from historical behaviors w.r.t. given ads, which can improve the expressive ability of model greatly and better capture user's diverse interests.
%\item We develop two novel techniques to improve the performance of industrial deep neural networks. We develop a mini-batch aware regularizer to save the heavy computation of regularization on large scale parameters, which is helpful for avoiding overfitting. We design a data adaptive activation function, which generalizes PReLU by considering the distribution of inputs and shows well performance.
% \item We develop two novel techniques:  mini-batch aware regularization and data adaptive activation function, which is shown to be useful for training industrial deep networks with sparse inputs and hundreds of millions of parameters. 
%\item We deploy DIN in the display advertising system in Alibaba, which holds the largest market share of online advertising in China. DIN shows superior performance and contributes a significant improvement to the business.
%\end{itemize}

\begin{itemize}
\item We point out the limit of using fixed-length vector to express user's diverse interests and design a novel deep interest network (DIN) which introduces a local activation unit to adaptively learn the representation of user interests from historical behaviors w.r.t. given ads. DIN can improve the expressive ability of model greatly and better capture the diversity characteristic of user interests. 
\item We develop two novel techniques to help training industrial deep networks: i) a mini-batch aware regularizer, which saves heavy computation of regularization on deep networks with huge number of parameters and is helpful for avoiding overfitting, ii) a data adaptive activation function, which generalizes PReLU by considering the distribution of inputs and shows well performance.
\item We conduct extensive experiments on both public and Alibaba datasets. Results verify the effectiveness of proposed DIN and training techniques. Our code\footnote{Experiment code on two public datasets is available on GitHub: https://github.com/zhougr1993/DeepInterestNetwork\label{code}} is publicly available. The proposed approaches have been deployed in the commercial display advertising system in Alibaba, one of world's largest advertising platform, contributing significant improvement to the business. 
%\item deploy DIN in the display advertising system in Alibaba, which holds the largest market share of online advertising in China. DIN shows superior performance and contributes a significant improvement to the business.
\end{itemize}

%\begin{itemize}
%\item We propose a deep interest network (DIN) which designs an activation unit to adaptively learn the representation of user interests from historical behaviors w.r.t. given ads. It can better capture the characteristic of user behavior data and improve the expressive ability of model greatly.% under limited embedding dimension.
%\item We develop two novel techniques:  mini-batch aware regularization and data adaptive activation function, which is shown to be useful for training industrial deep networks with sparse inputs and hundreds of millions of parameters. 
%\end{itemize}

In this paper we focus on the CTR prediction modeling in the scenario of display advertising in e-commerce industry. 
Methods discussed here can be applied in similar scenarios with rich user behaviors, such as personalized recommendation in e-commerce sites, feeds ranking in social networks etc.

The rest of the paper is organized as follows. We discuss related work in section 2 and introduce the background about characteristic of user behavior data in display advertising system of e-commerce site in section 3. Section 4 and 5 describe in detail the design of DIN model as well as two proposed training techniques. We present experiments in section 6 and conclude in section 7. 

%% file: ch_relatedwork.tex
\section{Relatedwork}
The structure of CTR prediction model has evolved from shallow to deep. At the same time, the number of samples and the dimension of features used in CTR model have become larger and larger. In order to better extract feature relations to improve performance, several works pay attention to the design of model structure.

As a pioneer work, NNLM \cite{bengio:nnlm} learns distributed representation for each word,
aiming to avoid curse of dimension in language modeling.
This method, often referred to as embedding,
has inspired many natural language models and CTR prediction models that need to handle large-scale sparse inputs.

LS-PLM \cite{MLR} and FM \cite{rendle:fm} models can be viewed as a class of networks with one hidden layer, which first 
 employs embedding layer on sparse inputs and then imposes specially designed transformation functions for target fitting, aiming to capture the combination relations among features.

Deep Crossing \cite{deep_crossing}, Wide\&Deep Learning \cite{widedeep} and YouTube Recommendation CTR model \cite{youtube:recommend} extend LS-PLM and FM by replacing the transformation function with complex MLP network, which enhances the model capability greatly. PNN\cite{PNN} tries to capture high-order feature interactions by involving a product layer after embedding layer. DeepFM\cite{DeepFM} imposes a factorization machines as "wide" module in Wide\&Deep \cite{widedeep} with no need of feature engineering.
Overall, these methods follow a similar model structure with combination of embedding layer (for learning the dense representation of sparse features) and MLP (for learning the combination relations of features automatically).
This kind of CTR prediction model reduces the manual feature engineering jobs greatly. %to a great extent.
Our base model follows this kind of model structure. 
However in applications with rich user behaviors, features are often contained with variable-length list of ids, e.g., searched terms or watched videos in YouTube recommender system \cite{youtube:recommend}. These models often transform corresponding list of embedding vectors into a fixed-length vector via sum/average pooling, which causes loss of information.
The proposed DIN tackles it by adaptively learning the representation vector w.r.t. given ad, improving the expressive ability of model.   

Attention mechanism originates from Neural Machine Translation (NMT) field \cite{bengio:attention}.
NMT takes a weighted sum of all the annotations to get an expected annotation and focuses only on information relevant to the generation of next target word.
A recent work, DeepIntent \cite{deep_intent} applies attention in the context of search advertising. Similar to NMT, they use RNN\cite{rnn} to model text, then learn one global hidden vector to help paying attention on the key words in each query. It is shown that the use of attention can help capturing the main intent of query or ad.
DIN designs a local activation unit to soft-search for relevant user behaviors and takes a weighted sum pooling to obtain the adaptive representation of user interests with respect to a given ad. The user representation vector varies over different ads, which is different from DeepIntent in which there is no interaction between ad and user.

We make code publicly available, and further show how to successfully deploy DIN in one of the world's largest advertising systems with novel developed techniques for training large scale deep networks with hundreds of millions of parameters. 

%% file: ch_sysview.tex
%\section{Overview of Display Advertising System}
\section{Background}
\label{sec:bg}

\begin{figure}
\centering
\includegraphics[height=2.5in, width=2.5in,keepaspectratio]{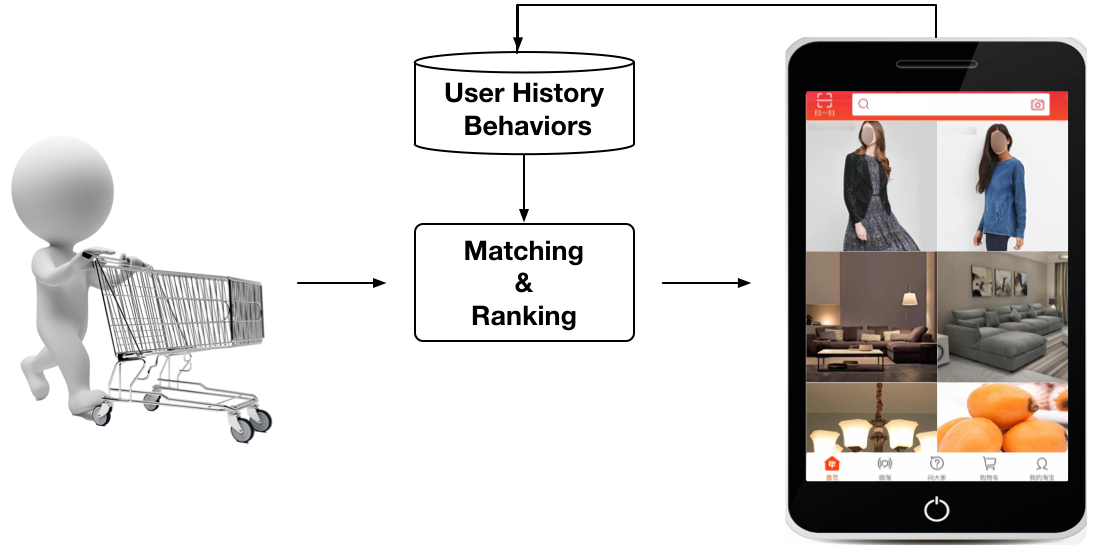}
\caption{Illustration of running procedure of display advertising system in Alibaba, in which user behavior data plays important roles.}
\label{figure_display_ad_scenario}
\vspace{-0.4cm}
\end{figure}

%The overall scenario of the display advertising system is illustrated in Figure   \ref{figure_display_ad_scenario}.
In e-commerce sites, such as Alibaba, advertisements are natural goods. In the rest of this paper, without special declaration, we regard ads as goods. Figure \ref{figure_display_ad_scenario} briefly illustrates the running procedure of display advertising system in Alibaba, which consists of two main stages: i) matching stage which generates list of candidate ads relevant to the visiting user via methods like collaborative filtering, ii) ranking stage which predicts CTR for each given ad and then selects top ranked ones.    
%When a user visits the e-commerce site, system
%i) checks his/her historical behavior data;
%ii) generates candidate ads by matching module;
%iii) predicts the click probability (CTR) of each ad and displays top ranked ads under certain ranking score function;
%iv) logs user reactions (click or not) given the displayed ads.
Everyday, hundreds of millions of users visit the e-commerce site, leaving us with lots of user behavior data which contributes critically in building matching and ranking models.
%Rich user historical behavior data contains hint about user interests. 
It is worth mentioning that users with rich historical behaviors contain diverse interests. 
For example, a young mother has browsed goods including woolen coat, T-shits, earrings, tote bag, leather handbag and children's coat recently. These behavior data give us hints about her shopping interests.
When she visits the e-commerce site, system displays a suitable ad to her, for example a new handbag.
Obviously the displayed ad only matches or activates part of interests of this mother. 
In summary, interests of user with rich behaviors are \textsl{\textbf{diverse}} and could be \textsl{\textbf{locally activated}} given certain ads. We show later in this paper making use of these characteristics plays important role for building CTR prediction model.

%% file: ch_approach.tex
%Our click model is designed for CTR prediction of display advertising. The observations of this application are Users' profile information, Ads' static attributes, and Users' behavior history. Different from the sponsored search, users come into display advertising system without clear target. Thus we wish to mining users' interest from the context information and estimate the CTR of each display accurately.
\section{Deep Interest Network}\label{sec:opa}

Different from sponsored search, users come into display advertising system without explicitly expressed intentions. 
Effective approaches are required to extract user interests from rich historical behaviors when building the CTR prediction model. 
Features that depict users and ads are the basic elements in the CTR modeling of advertisement system. 
Making use of these features reasonably and mining information from them are critical.

\subsection{Feature Representation}
Data in industrial CTR prediction tasks is mostly in a multi-group categorial form, for example, [weekday=Friday, gender=Female, \\ visited$\_$cate$\_$ids=$\{$Bag,Book$\}$, ad$\_$cate$\_$id=Book],
which is normally transformed into high-dimensional sparse binary features via encoding \cite{deep_crossing,widedeep,ftrl}. 
Mathematically, encoding vector of i-th feature group is formularized as $\textbf{t}_i \in R^{K_i}$. $K_i$ denotes the dimensionality of feature group $i$, which means feature group $i$ contains $K_i$ unique ids. 
$\textbf{t}_i[j]$ is the j-th element of $\textbf{t}_i$  and $\textbf{t}_i[j] \in \{0,1\}$. $\sum_{j=1}^{K_i} \textbf{t}_i[j]=k$. 
Vector $\textbf{t}_i$ with $k=1$ refers to one-hot encoding and $k>1$ refers to multi-hot encoding. 
Then one instance can be represent as $\bs{x} = [\bs{t}_1^T, \bs{t}_2^T,...\bs{t}_M^T]^T$ in a group-wise manner, where $M$ is number of feature groups, $\sum_{i=1}^{M}K_i = K$, $K$ is dimensionality of the entire feature space. 
In this way, the aforementioned instance with four groups of features are illustrated as: 
%\begin{equation}
\[
%  \underbrace{a + \overbrace{b+\cdots}^{{}=t}+z}_{\text{total}} ~~ a + {\overbrace{b+\cdots}^{126}+z
\underbrace{[0,0,0,0,1,0,0]}_{\text{weekday=Friday}} ~ \underbrace{[0,1]}_{\text{gender=Female}} ~ \underbrace{[0,..,1,...,1,...0]}_{\text{visited\_cate\_ids=\{Bag,Book\}}} 
~ \underbrace{[0,..,1,...,0]}_{\text{ad\_cate\_id=Book}}
\]
%\end{equation}

The whole feature set used in our system is described in Table \ref{table_feature_set}.
It is composed of four categories, among which user behavior features are typically multi-hot encoding vectors and contain rich information of user interests. 
Note that in our setting, there are no combination features. We capture the interaction of features with deep neural network. 

\begin{table}
\caption{Statistics of feature sets used in the display advertising system in Alibaba. Features are composed of sparse binary vectors in the group-wise manner.} % The dimensionality of fine-grained $goods\_ids$ feature scales up to $10^9$.}
\centering
\resizebox{0.47\textwidth}{!}{%
\label{table:fea-table}
\begin{tabular}{llccc}
\toprule
Category & Feature Group & Dimemsionality & Type & \#Nonzero Ids per Instance\\ \toprule
\multirow{3}{7em}{User Profile Features }
& gender & 2 &  one-hot & 1  \\ \cline{2-5}
& age\_level & $\sim 10$ &  one-hot & 1  \\ \cline{2-5}
& ... & ... & ... & ...  \\ \midrule

\multirow{4}{7em}{User Behavior Features}
& visited\_goods\_ids& $\sim 10^9$ &  multi-hot & $\sim 10^3$  \\ \cline{2-5}
& visited\_shop\_ids& $\sim 10^7$ &  multi-hot & $\sim 10^3$  \\ \cline{2-5}
& visited\_cate\_ids& $\sim 10^4$ &  multi-hot & $\sim 10^2$  \\\midrule

\multirow{4}{7em}{Ad Features }
& goods\_id & $\sim 10^7$ &  one-hot & 1  \\ \cline{2-5}
& shop\_id & $\sim 10^5$ &  one-hot & 1  \\ \cline{2-5}
& cate\_id & $\sim 10^4$ &  one-hot & 1   \\ \cline{2-5}
& ... & ... & ... & ...  \\ \midrule

\multirow{3}{7em}{Context Features }
& pid & $\sim 10$ &  one-hot & 1  \\ \cline{2-5}
& time & $\sim 10$ &  one-hot & 1  \\ \cline{2-5}
& ... & ... & ... & ...  \\ \bottomrule
\end{tabular}}
\label{table_feature_set}
\end{table}

\begin{figure*}[!t]
\centering
\includegraphics[height=7in, width=7in,keepaspectratio]{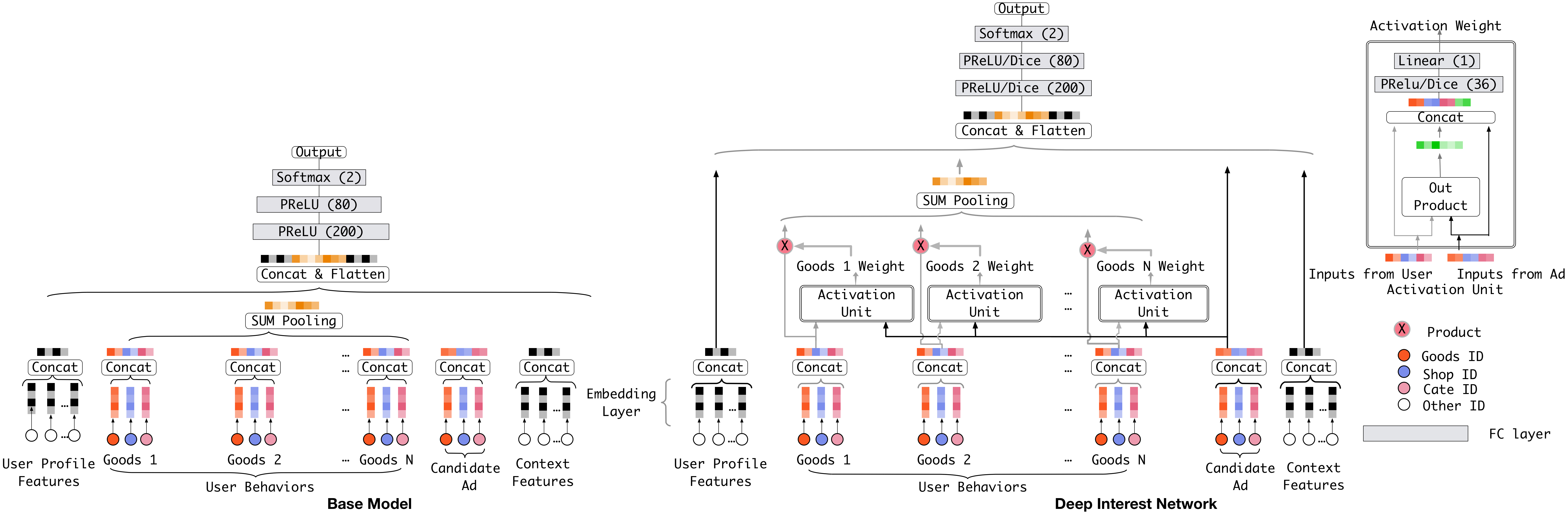}
\caption{Network Architecture. The left part illustrates the network of base model (Embedding\&MLP). Embeddings of cate\_id, shop\_id and goods\_id belong to one goods are concatenated to represent one visited goods in user's behaviors. Right part is our proposed DIN model. It introduces a local activation unit, with which the representation of user interests varies adaptively given different candidate ads.}
\label{model_arch}
\end{figure*}

\subsection{Base Model(Embedding\&MLP)}
\label{sec:basemodel}
%Following the popular model structure introduced in \cite{deep_crossing,widedeep,youtube:recommend}, we design our base model as shown in the left of Fig.\ref{model_arch}.
Most of the popular model structures \cite{deep_crossing,widedeep,youtube:recommend} share a similar Embedding\&MLP paradigm, which we refer to as base model, as shown in the left of Fig.\ref{model_arch}. It consists of several parts:
%It follows the embedding\&MLP architecture. %, with all the parameters trained in the end-to-end manner.
\paragraph{\textbf{Embedding layer}} As the inputs are high dimensional binary vectors, embedding layer is used to transform them into low dimensional dense representations. 
%Let $\bs{t}_i\in \mathbb{R}^{K_i}$ represent the feature vector of i-th group in Table \ref{table_feature_set}. 
%Then one instance can be represent as $\bs{x} = [\bs{t}_1^T, \bs{t}_2^T,...\bs{t}_M^T]^T$ in a group-wise manner, where $M$ is number of feature groups, $\sum_{i=1}^{M}K_i = K$, $K$ is dimensionality of the entire feature space. 
For the i-th feature group of $\bs{t}_i$,  let $\mathrm{W}^i = [w_1^i,...,w_j^i, ..., w_{K_i}^i] \in\mathbb{R}^{D \times K_i}$ represent the i-th embedding dictionary, where $w_j^i \in R^D$ is an embedding vector with dimensionality of $D$.  
%Given instance of $\bs{x} = [\bs{t}_1^T, \bs{t}_2^T,...\bs{t}_M^T]^T$.
Embedding operation follows the table lookup mechanism, as illustrated in Fig.\ref{model_arch}. 
\begin{itemize}
\item If $\bs{t}_i$ is one-hot vector with j-th element $\bs{t}_i[j]=1$, the embedded representation of $\bs{t}_i$ is a single embedding vector $\bs{e}_i = w_j^i$. 
\item If $\bs{t}_i$ is multi-hot vector with $\bs{t}_i[j]=1 ~\text{for}~ j \in \{i_1, i_2, ..., i_k\} $, the embedded representation of $\bs{t}_i$ is a list of embedding vectors: $\{\bs{e}_{i_1}, \bs{e}_{i_2},...\bs{e}_{i_k}\} = \{w_{i_1}^i, w_{i_2}^i, ...w_{i_k}^i\}$.
\end{itemize}

\paragraph{\textbf{Pooling layer and Concat layer}} 
Notice that different users have different numbers of behaviors. Thus the number of non-zero values for multi-hot behavioral feature vector $\bs{t}_i$  varies across instances, causing the lengths of the corresponding list of embedding vectors to be variable.
As fully connected networks can only handle fixed-length inputs, it is a common practice \cite{widedeep,youtube:recommend} to transform the list of embedding vectors via a pooling layer to get a fixed-length vector: 
\begin{equation}
\label{eq:pooling}
\bs{e}_i = \text{pooling}(\bs{e}_{i_1}, \bs{e}_{i_2},...\bs{e}_{i_k}).	
\end{equation}
Two most commonly used pooling layers are sum pooling and average pooling, which apply element-wise sum/average operations to the list of embedding vectors. 

Both embedding and pooling layers operate in a group-wise manner, mapping the original sparse features into multiple fixed-length representation vectors.   
Then all the vectors are concatenated together to obtain the overall representation vector for the instance. 

\paragraph{\textbf{MLP}} Given the concatenated dense representation vector, fully connected layers are used to learn the combination of features automatically. 
Recently developed methods \cite{widedeep,DeepFM,PNN} focus on designing structures of MLP for better information extraction. \par
\paragraph{\textbf{Loss}}The objective function used in base model is the negative log-likelihood function defined as:
\begin{equation}
    L= - \frac{1}{N} \sum_{(\bs{x},y)\in\mathcal{S}} (y\log p(\bs{x}) + (1-y)\log(1-p(\bs{x}))),
\end{equation}
where $\mathcal{S}$ is the training set of size $N$, with $\bs{x}$ as the input of the network and $y\in \{0,1\}$ as the label, $p(\bs{x})$ is the output of the network after the softmax layer, representing the predicted probability of sample $\bs{x}$ being clicked.

\subsection{The structure of Deep Interest Network}

Among all those features of Table \ref{table_feature_set}, user behavior features are critically important and play key roles in modeling user interests in the scenario of e-commerce applications.   

Base model obtains a fixed-length representation vector of user interests by pooling all the embedding vectors over the user behavior feature group, as Eq.(\ref{eq:pooling}). This representation vector stays the same for a given user, in regardless of what candidate ads are.
In this way, the user representation vector with a limited dimension will be a bottleneck to express user's diverse interests.
To make it capable enough, an easy method is to expand the dimension of embedding vector, which unfortunately will increase the size of learning parameters heavily.
It will lead to overfitting under limited training data and add the burden of computation and storage, which may not be tolerated for an industrial online system. 

%As discussed above, when visiting e-commerce sites, users are often interested in many kinds of goods. 
%In other words, user interests are diverse.  
%However, it is known that users are often with diverse interests when visiting e-commerce sites. 
%Base model obtains the user interest vector by average pooling over the embedding of all behaviors. Thus base model obtains a same interest vector of users given different candidate ads for one user.
% In the base model, sums the embedding of all user behavior ids to obtain the user interest vector. The model obtains the same user % interest vector when predicts different candidate ad for one user.
%In order to embed diverse user interests in this vector, one easy method is expanding the dimension to make it capable enough to contain information of different interests. 
%Unfortunately, high dimension of embedding vectors will increase the size of learning parameters heavily.
%It will lead to overfitting under limited training data and add the burden of computation and storage, which may not be tolerated for an industrial online system.

Is there an elegant way to represent user's diverse interests in one vector under limited dimension?
The local activation characteristic of user interests gives us inspiration to design a novel model named deep interest network(DIN).
Imagine when the young mother mentioned above in section \ref{sec:bg} visits the e-commerce site, she finds the displayed new handbag cute and clicks it.
Let's dissect the driving force of click action. 
The displayed ad hits the related interests of this young mother by soft-searching her historical behaviors and finding that she had browsed similar goods of tote bag and leather handbag recently.  
In other words, behaviors related to displayed ad greatly contribute to the click action.  
DIN simulates this process by paying attention to the representation of locally activated interests w.r.t. given ad.
Instead of expressing all user's diverse interests with the same vector, DIN adaptively calculate the representation vector of user interests by taking into consideration the relevance of historical behaviors w.r.t. candidate ad.
This representation vector varies over different ads.         

The right part of Fig.\ref{model_arch} illustrates the architecture of DIN. 
Compared with base model, DIN introduces a novel designed local activation unit and maintains the other structures the same.  Specifically, activation units are applied on the user behavior features, which performs as a weighted sum pooling to adaptively calculate user representation $\bs{v}_U$ given a candidate ad $A$, as shown in Eq.(\ref{eq:attention})

%By simulating this process, we   
%Instead of focusing on a global representation vector, we actually only need to pay attention to the representation of locally related interests given a candidate ad.  
%In this way, representation vector of user interest is calculated adaptively given $\prec$user, ad$\succ$ pair. 
%That is, it varies over different candidate ads.  
%Then user vector only needs to represent related interest when faced different candidate ad, which is within the ability of a vector with appropriate dimension. \par
%For different candidate ads, we want our model to find related part of interests automatically. \par
%This candidate specific interest vector can be realized by attention mechanism\cite{bengio:attention}, which derive from neural machine translation.
%Under this consideration, we design a novel model named deep interest network(DIN), to better capture the characteristic of user behavior data and help improving the expressive ability of model under limited embedding dimension. 
%It is illustrated in the right of Fig.\ref{model_arch}.
%Compared with base model, DIN introduces a new designed activation unit and maintains the other net structures the same.  

%Specifically, activation unit is applied on the user behavior feature groups, which performs as a weighted average pooling to adaptively calculate user representation $\bs{v}_U$ given a candidate ad $A$, as shown in Eq.(\ref{eq:attention})
{
\begin{small}
\begin{eqnarray}
\begin{split}
\label{eq:attention}
\bs{v}_U(A) = f(\bs{v}_A,\bs{e}_1,\bs{e}_2,..,\bs{e}_H) & = \sum_{j=1}^H a(\bs{e}_j,\bs{v}_A) \bs{e}_j = \sum_{j=1}^H \bs{w}_j \bs{e}_j,
\end{split}
\end{eqnarray}
\end{small}
}
where $\{\bs{e}_1, \bs{e}_2, ..., \bs{e}_H\}$ is the list of embedding vectors of behaviors of user $U$ with length of $H$, $\bs{v}_A$ is the embedding vector of ad $A$. In this way, $\bs{v}_U(A)$ varies over different ads. $a(\cdot)$ is a feed-forward network with output as the activation weight, as illustrated in Fig.\ref{model_arch}. Apart from the two input embedding vectors, $a(\cdot)$ adds the out product of them to feed into the subsequent network, which is an explicit knowledge to help relevance   modeling.   

Local activation unit of Eq.(\ref{eq:attention}) shares similar ideas with attention methods which are developed in NMT task\cite{bengio:attention}.
However different from traditional attention method, the constraint of $\sum_i w_i = 1$ is relaxed in Eq.(\ref{eq:attention}), aiming to reserve the intensity of user interests. 
That is, normalization with softmax on the output of $a(\cdot)$ is abandoned.
Instead, value of $\sum_i w_i$ is treated as an approximation of the intensity of activated user interests to some degree.   
For example, if one user's historical behaviors contain 90\% clothes and 10\% electronics. 
Given two candidate ads of T-shirt and phone, T-shirt activates most of the historical behaviors belonging to clothes and may get larger value of $\bs{v}_U$ (higher intensity of interest) than phone. 
Traditional attention methods lose the resolution on the numerical scale of $\bs{v}_U$ by normalizing of the output of $a(\cdot)$. 
%Taking the original output of $a(\cdot)$  as activation score can be a simple way to maintain the intensity of interest, which helps making use of user historical behavior data more sufficiently.

%Assume user U is interested in clothes and electronics, with the intensity to be 90\% and 10\% relatively. 
%This causes the historical behaviors of clothes occur more \textbf{frequently} than electronics for $U$.     
%Assume two candidate ads T-shirt and iphone are selected as candidates for $U$.
%Value of representation vector $\bs{v}_U$ given T-shirt will be larger than given iphone.
%Indeed, taking the original output of $a(\cdot)$  as activation weight can be a simple way to maintain the intensity of interest, which helps us making use of user historical behavior more sufficiently.

We have tried LSTM to model user historical behavior data in the sequential manner.
But it shows no improvement. 
%But it makes little difference. 
Different from text which is under the constraint of grammar in NLP task, the sequence of user historical behaviors may contain multiple concurrent interests.
Rapid jumping and sudden ending over these interests causes the sequence data of user behaviors to seem to be noisy.
A possible direction is to design special structures to model such data in a sequence way.
We leave it for future research.

\section{Training Techniques}
In the advertising system in Alibaba, numbers of goods and users scale up to hundreds of millions.  
%There are billion kinds of goods in Alibaba, and our advertising system needs to serve more than hundred million users one day.
Practically, training industrial deep networks with large scale sparse input features is of great challenge.
In this section, we introduce two important techniques which are proven to be helpful in practice. 

\subsection{Mini-batch Aware Regularization}

%In order to capture more specific interests of users, we typically introduce billions of fine-grained ids to describe the users' behaviors. These sparse and high dimensional features may lead to severe overfitting problem.
%{\EDIT why these feature would cause overfitting?(1 more parameter ; 2 less times)}
%to capture user interest detailly. \par
%Not surprisingly, overfitting problem is encountered while training our model with large scale parameters and sparse inputs.
%In experiment, with addition of fine-grained user visited $good\_ids$ feature, model performance falls rapidly after the first epoch.\par
%It is known that internet-scale user behavior data follows the long-tail law, that is, lots of feature ids occur a few times in the training samples, while little of them occur many times.
%This inevitably introduces noise into the training process and intensifies overfitting.
%An easy way to reduce overfitting is to filter out those low-frequency feature ids, which can be viewed as manual regularization.
%However, such frequency based filter is quite rough in terms of information loss and threshold setting.
%Many methods have been proposed to reduce overfitting, such as $\ell_2$ and $\ell_1$ regularization \cite{lasso}, and Dropout \cite{dropout}. 
%However, merely of them are suitable for training network with sparse inputs and billions of parameters.

%Practically overfitting problem is encountered while training industrial networks with large scale sparse input features.
Overfitting is a critical challenge for training industrial networks. 
For example, with addition of fine-grained features, such as features of $\textsl{goods\_ids}$ with dimensionality of 0.6 billion (including $visited\_goods\_ids$ of user and $goods\_id$ of ad as described in Table \ref{table_feature_set}), model performance falls rapidly after the first epoch during training without regularization, as the dark green line shown in Fig.\ref{fig:adaptive_reg} in later section \ref{sec:reg}.   
It is not practical to directly apply traditional regularization methods, such as $\ell_2$ and $\ell_1$ regularization, on training networks with sparse inputs and hundreds of millions of parameters.
Take $\ell_2$ regularization as an example.
Only parameters of non-zero sparse features appearing in each mini-batch needs to be updated in the scenario of SGD based optimization methods without regularization. 
However, when adding $\ell_2$ regularization it needs to calculate L2-norm over the whole parameters for each mini-batch, which leads to extremely heavy computations and is unacceptable with parameters scaling up to hundreds of millions.        
%Original $\ell_2$ regularization needs to calculated L2-norm over the whole parameters on each mini-batch, which leads to extremely heavy computations and is unacceptable in our situation. 

\begin{figure}[!ht]
\centering
\includegraphics[height=1.8in, width=2.2in,keepaspectratio]{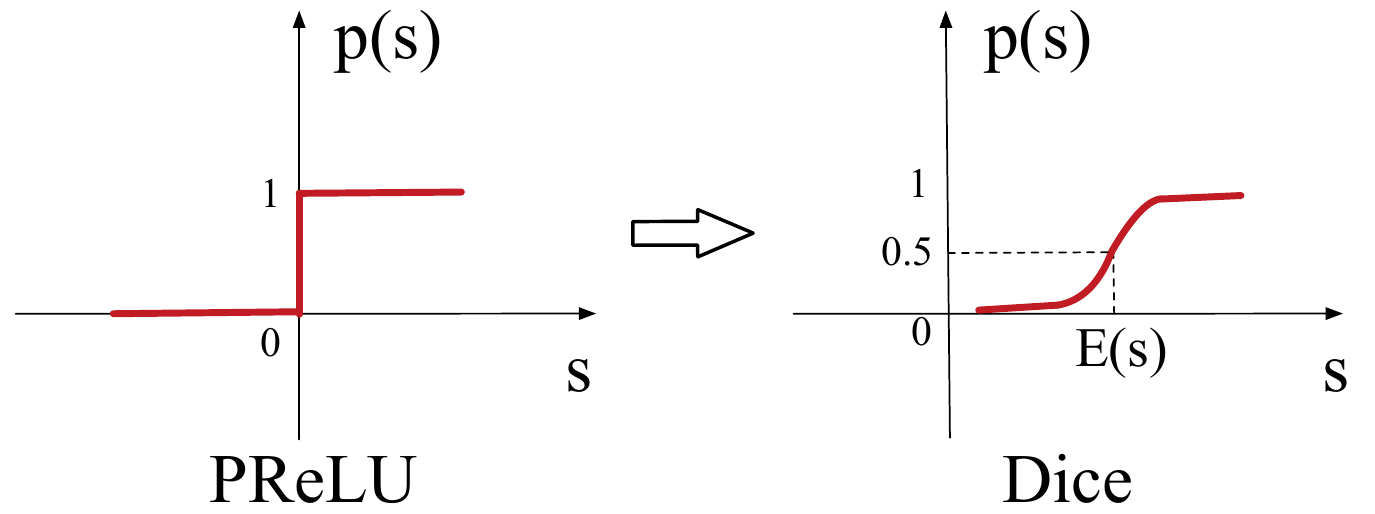}
\caption{Control function of PReLU and Dice.}
\label{figure_dice}
\vspace{-0.4cm}
\end{figure}

In this paper, we introduce an efficient mini-batch aware regularizer, which only calculates the L2-norm over the parameters of sparse features appearing in each mini-batch and makes the computation possible.     
In fact, it is the embedding dictionary that contributes most of the parameters for CTR networks and arises the difficulty of heavy computation. 
Let $\bs{\mathrm{W}}\in\mathbb{R}^{D\times K}$ denote parameters of the whole embedding dictionary, with $D$ as the dimensionality of the embedding vector and $K$ as the dimensionality of feature space. 
Expand the $\ell_2$ regularization on $\bs{\mathrm{W}}$ over samples 
%\begin{equation}
%\label{eq:batch}
%\sum_{i=1}^{n_{dj}} \frac{I_{id}}{n_{dj}}w_d^2
%\end{equation}
\begin{footnotesize}
\begin{eqnarray}
\begin{split}
\label{eq:batch}
L_2(\bs{\mathrm{W}}) &= \|\bs{\mathrm{W}}\|_2^2 = \sum_{j=1}^{K}\|\bs{w}_j\|_2^2 = \sum_{(\bs{x},y)\in\mathcal{S}}\sum_{j=1}^{K} \frac{I(\bs{x}_j\neq0)}{n_{j}}\|\bs{w}_j\|_2^2,
\end{split}
\end{eqnarray}
\end{footnotesize}
where $\bs{w}_j\in\mathbb{R}^{D}$ is the $j$-th embedding vector,
%which corresponding to the number of feature IDs in our applications.
$I(\bs{x}_j\neq0)$ denotes if the instance $\bs{x}$ has the feature id $j$, and $n_j$ denotes the number of occurrence for feature id $j$ in all samples. %Here we use $I(\cdot)$ as the indicator function.
%When applying the gradient descent optimization on mini-batches, the regularization is transformed into
Eq.(\ref{eq:batch}) can be transformed into Eq.(\ref{eq:minibatch}) in the mini-batch aware manner 
\begin{small}
\begin{equation}
\label{eq:minibatch}
L_2(\bs{\mathrm{W}}) = \sum_{j=1}^{K}\sum_{m=1}^{B}\sum_{(\bs{x},y)\in\mathcal{B}_m}\frac{I(\bs{x}_j\neq 0)}{n_{j}}\|\bs{w}_j\|_2^2,
\end{equation}
\end{small}
where $B$ denotes the number of mini-batches, $\mathcal{B}_m$ denotes the $m$-th mini-batch.
%To simplify the computation, 
Let $\alpha_{mj} = \max_{(\bs{x},y)\in\mathcal{B}_m} I(\bs{x}_j\neq0)$
%{\small
%\begin{equation}
%\alpha_{mj} = \max_{(\bs{x},y)\in\mathcal{B}_m} I(\bs{x}_j\neq0),
%\end{equation}}
denote if there is at least one instance having the feature id $j$ in mini-batch $\mathcal{B}_m$.
Then Eq.(\ref{eq:minibatch}) can be approximated by
\begin{small}
\begin{equation}
\label{eq:approx}
L_2(\bs{\mathrm{W}})\approx\sum_{j=1}^{K} \sum_{m=1}^{B}  \frac{\alpha_{mj}}{n_j}\|\bs{w}_j\|_2^2.
\end{equation}
\end{small}
In this way, we derive an approximated mini-batch aware version of $\ell_2$ regularization.
For the $m$-th mini-batch, the gradient w.r.t. the embedding weights of feature $j$ is
%\begin{equation}	
%\scalebox{0.9}{
%\parbox{0.37\textwidth}{
%\begin{eqnarray}
%\begin{split}
%\bs{w}_j \leftarrow \bs{w}_j- \eta
%    \left[
%       \frac{1}{|\mathcal{B}_m|} \sum_{(\bs{x},y) \in \mathcal{B}_m} { \frac{\partial L(p(\bs{x}), y)}{\partial \bs{w}_j}
%       +\lambda \frac{\alpha_{mj}}{n_j}\bs{w}_j}
%    \right].
%\end{split}
%\end{eqnarray}
%}
%}
%\end{equation}
\begin{small}
\begin{equation}
\bs{w}_j \leftarrow \bs{w}_j- \eta
    \left[
       \frac{1}{|\mathcal{B}_m|} \sum_{(\bs{x},y) \in \mathcal{B}_m} { \frac{\partial L(p(\bs{x}), y)}{\partial \bs{w}_j}
       +\lambda \frac{\alpha_{mj}}{n_j}\bs{w}_j}
    \right],
\end{equation}
\end{small}
in which only parameters of features appearing in $m$-th mini-batch participate in the computation of regularization.

\subsection{Data Adaptive Activation Function}
%{\EDIT
%In order to make DIN to be more suitable for industrial-scale net sparse input, besides the design of net structure and choice of input, we propose a new activation function to accelerate the convergence.

%Industrial-scale dimensionality of sparse features and data size give challenges to the convergence for training deep networks. 
%To fit the data better, we propose a data adaptive activation function to help convergence.

PReLU \cite{ms:prelu} is a commonly used activation function
\begin{small}
\begin{equation}
f(s) = \begin{cases}
       s & \text{ if } s > 0  \\
       \alpha s & \text{ if } s \leq 0.
       \end{cases}
     = ~ p(s) \cdot s + (1-p(s))\cdot \alpha s,   
\end{equation}
\end{small}
where $s$ is one dimension of the input of activation function $f(\cdot)$ and $p(s) = I(s>0)$ is an indicator function which controls $f(s)$ to switch between two channels of $f(s)=s$ and $f(s)=\alpha s$. $\alpha$ in the second channel is a learning parameter. 
Here we refer to $p(s)$ as the control function. The left part of Fig.\ref{figure_dice} plots the control function of PReLU.
PReLU takes a hard rectified point with value of $0$, which may be not suitable when the inputs of each layer follow different distributions. 
Take this into consideration, we design a novel data adaptive activation function named \textsl{\textbf{Dice}},
\begin{equation}
f(s) = p(s) \cdot s + (1-p(s))\cdot \alpha s, \ \ p(s) = \frac{1}{ 1 + e^{- \frac{s - E[s]}{\sqrt{Var[s] + \epsilon}}}}
\end{equation}
with the control function to be plotted in the right part of Fig.\ref{figure_dice}.
In the training phrase, $E[s]$ and $Var[s]$ is the mean and variance of input in each mini-batch.  
In the testing phrase, $E[s]$ and $Var[s]$ is calculated by moving averages $E[s]$ and $Var[s]$ over data. 
$\epsilon$ is a small constant which is set to be $10^{-8}$ in our practice.

Dice can be viewed as a generalization of PReLu.
The key idea of Dice is to adaptively adjust the rectified point according to distribution of input data, whose value is set to be the mean of input.
Besides, Dice controls smoothly to switch between the two channels. 
When $E(s)=0$ and $Var[s]=0$, Dice degenerates into PReLU.

%% file: ch_exp.tex
\section{Experiments} \label{exp}
%In this section, We first compare DIN with other methods including Wide\&Deep\cite{widedeep}, PNN\cite{PNN}, deepFM\cite{DeepFM} and Logistic Regression on two public dataset. DINs need to be evaluated on dataset with user behaviors, but not all dataset about CTR prediction contains user behaviors. Two public datasets with user behaviors: Amazon Dataset\footnote{http://jmcauley.ucsd.edu/data/amazon/} and MovieLens\footnote{https://grouplens.org/datasets/movielens/20m/} are used in this paper. Then our proposed methods are evaluated on Alibaba product data. We evaluate the effectiveness of the proposed structure of Deep Interest Network, activation function: Dice as well as the Mini-batch Aware Regularization Technique.

%In this section, we conduct empirical study on the performance of proposed DIN model and training techniques.
In this section, we present our experiments in detail, including datasets, evaluation metric, experimental setup, model comparison and the corresponding analysis.
Experiments on two public datasets with user behaviors as well as a dataset collected from the display advertising system in Alibaba demonstrate the effectiveness of proposed approach which outperforms state-of-the-art methods on the CTR prediction task.
Both the public datasets and experiment codes are made available\textsuperscript{\ref{code}}.%\footnote{Experiment codes on two public datasets are available at GitHub: https://github.com/zhougr1993/DeepInterestNetwork}.

\subsection{Datasets and Experimental Setup}
\textbf{Amazon Dataset\footnote{http://jmcauley.ucsd.edu/data/amazon/}.} 
Amazon Dataset contains product reviews and metadata from Amazon, which is used as benchmark dataset\cite{Amazon:AUC,AmazonData,ICCV_Amazon}. We conduct experiments on a subset named Electronics, which contains 192,403 users, 63,001 goods, 801 categories and 1,689,188 samples. User behaviors in this dataset are rich, with more than 5 reviews for each users and goods. Features include goods\_id, cate\_id, user reviewed goods\_id\_list and cate\_id\_list. Let all behaviors of a user be ($b_1, b_2, \ldots, b_k, \ldots, b_n$), the task is to predict the (k+1)-th reviewed goods by making use of the first k reviewed goods. 
    Training dataset is generated with $k = 1, 2, \ldots, n - 2$ for each user. In the test set, we predict the last one given the first $n-1$ reviewed goods. 
    For all models, we use SGD as the optimizer with exponential decay, in which learning rate starts at 1 and decay rate is set to 0.1. %We also tried Adam\cite{Adam}, which provides faster convergency but brings worse results than SGD for all deep models. 
    The mini-batch size is set to be $32$.

\textbf{MovieLens Dataset\footnote{https://grouplens.org/datasets/movielens/20m/}.} MovieLens data\cite{MovieLens} contains 138,493 users, 27,278 movies, 21 categories and 20,000,263 samples. To make it suitable for CTR prediction task, we transform it into a binary classification data. Original user rating of the movies is continuous value ranging from 0 to 5. We label the samples with rating of 4 and 5 to be positive and the rest to be negative. We segment the data into training and testing dataset based on userID. Among all 138,493 users, of which 100,000 are randomly selected into training set (about 14,470,000 samples) and the rest 38,493 into the test set (about 5,530,000 samples). The task is to predict whether user will rate a given movie to be above 3(positive label) based on historical behaviors. Features include movie\_id, movie\_cate\_id and user rated movie\_id\_list, movie\_cate\_id\_list. We use the same optimizer, learning rate and mini-batch size as described on Amazon Dataset.

\textbf{Alibaba Dataset.} We collected traffic logs from the online display advertising system in Alibaba, of which two weeks' samples are used for training and  samples of the following day for testing. The size of training and testing set is about 2 billions and 0.14 billion respectively. For all the deep models, the dimensionality of embedding vector is 12 for the whole 16 groups of features. %So the input dimension of first layer in MLP is 192. 
Layers of MLP is set by $192 \times 200 \times 80 \times 2$. Due to the huge size of data, we set the mini-batch size to be 5000 and use Adam\cite{Adam} as the optimizer. We apply exponential decay, in which learning rate starts at 0.001 and decay rate is set to 0.9.

The statistics of all the above datasets is shown in Table \ref{table:Statistics}. 
Volume of Alibaba Dataset is much larger than both Amazon and MovieLens, which brings more challenges.

\begin{table}[]
\caption{Statistics of datasets used in this paper.}
\small
\centering
\begin{threeparttable}
\begin{tabular}{lcccc}
\toprule
    Dataset       & Users & Goods\tnote{a} & Categories & Samples \\ \midrule
Amazon(Electro). & 192,403 & 63,001 & 801 & 1,689,188 \\ 
MovieLens. & 138,493 & 27,278 & 21 & 20,000,263 \\
Alibaba.  & 60 million & 0.6 billion & 100,000 & 2.14 billion \\ 
\bottomrule
\end{tabular}
\begin{tablenotes}%[para,flushleft]
        \item[a] For MovieLens dataset, goods refer to be movies.
        %\item[2] For security reasons we can only give an approximate number.
      \end{tablenotes}
      \end{threeparttable}
\label{table:Statistics}
\end{table}

\begin{figure*}[!t]
\centering
\includegraphics[height=6.2in, width=6.2in, keepaspectratio]{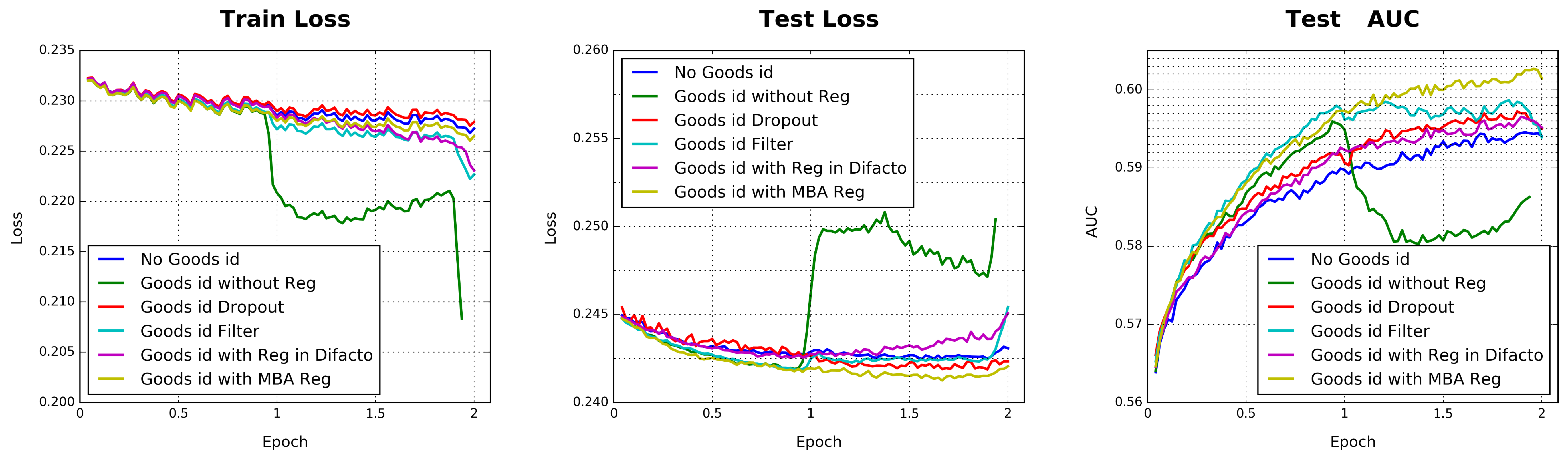}
\caption{Performances of BaseModel with different regularizations on Alibaba Dataset. Training with fine-grained $goods\_ids$ features without regularization encounters serious overfitting after the first epoch. All the regularizations show improvement, among which our proposed mini-batch aware regularization performs best. Besides, well trained model with $goods\_ids$ features gets higher AUC than without them. It comes from the richer information that fine-grained features contained.}
\label{fig:adaptive_reg}
\end{figure*}

\subsection{Competitors}
\label{competitors}
\begin{itemize}
\item \textbf{LR\cite{ftrl}}. Logistic regression (LR) is a widely used shallow model before deep networks for CTR prediction task. We implement it as a weak baseline.       
\item \textbf{BaseModel}. As introduced in section\ref{sec:basemodel}, BaseModel follows the Embedding\&MLP architecture and is the base of most of subsequently developed deep networks for CTR modeling. It acts as a strong baseline for our model comparison.   
\item \textbf{Wide\&Deep\cite{widedeep}}. 
In real industrial applications, Wide\&Deep model has been widely accepted. 
It consists of two parts: i) wide model, which handles the manually designed cross product features, ii) deep model, which automatically extracts nonlinear relations among features and equals to the BaseModel. 
Wide\&Deep needs expertise feature engineering on the input of the "wide" module. 
We follow the practice in \cite{DeepFM} to take cross-product of user behaviors and candidates as wide inputs.
For example, in MovieLens dataset, it refers to the cross-product of user rated movies and candidate movies. 
\item \textbf{PNN\cite{PNN}.} PNN can be viewed as an improved version of BaseModel by introducing a product layer after embedding layer to capture high-order feature interactions. 
\item \textbf{DeepFM\cite{DeepFM}}. It imposes a factorization machines as "wide" module in Wide\&Deep saving feature engineering jobs. 
\end{itemize}
\subsection{Metrics}
\label{metric}
In CTR prediction field, AUC is a widely used metric\cite{fawcett:roc}. 
It measures the goodness of order by ranking all the ads with predicted CTR, including intra-user and inter-user orders. 
An variation of user weighted AUC is introduced in \cite{Amazon:AUC,zhu2017optimized} which measures the goodness of intra-user order by averaging AUC over users and is shown to be more relevant to online performance in display advertising system. 
We adapt this metric in our experiments. For simplicity, we still refer it as AUC. 
It is calculated as follows:
\begin{small}
\begin{equation} \label{imps-guac}
\text{AUC} = \frac{\sum_{i=1}^{n} \#impression_i \times \text{AUC}_{i}}{\sum_{i=1}^{n} \#impression_i},
\end{equation}
\end{small}
where $n$ is the number of users, $\#impression_i$ and $\text{AUC}_i$ are the number of impressions and AUC corresponding to the $i$-th user. 

Besides, we follow \cite{yan2014coupled} to introduce RelaImpr metric to measure relative improvement over models. For a random guesser, the value of AUC is $0.5$. Hence RelaImpr is defined as below: 
\begin{small}
\begin{equation}
RelaImpr = \left(\frac{\text{AUC(measured model)} -0.5}{\text{AUC(base model)} - 0.5} - 1\right) \times 100\%.
\end{equation}
\end{small}

\begin{table}[]
\caption{Model Coparison on Amazon Dataset and MovieLens Dataset. All the lines calculate RelaImpr by comparing with BaseModel on each dataset respectively.}
\centering
\begin{threeparttable}
\begin{tabular}{lcccc}
\toprule
\multirow{2}{*}{Model}  & \multicolumn{2}{c}{MovieLens.} & \multicolumn{2}{c}{Amazon(Electro).}  \\ 
& AUC & RelaImpr & AUC & RelaImpr \\ \midrule
LR  & 0.7263 & -1.61\% & 0.7742 & -24.34\%  \\
BaseModel & 0.7300 & 0.00\% & 0.8624 & 0.00\%  \\
Wide\&Deep & 0.7304 & 0.17\% & 0.8637 & 0.36\%  \\
PNN  & 0.7321 & 0.91\% & 0.8679 & 1.52\%  \\
DeepFM  & 0.7324 & 1.04\% & 0.8683 & 1.63\% \\
\textbf{DIN} & \textbf{0.7337} & \textbf{1.61\%} & \textbf{0.8818} & \textbf{5.35}\%\\
\textbf{DIN with Dice}\tnote{a} & \textbf{0.7348} & \textbf{2.09\%} & \textbf{0.8871} & \textbf{6.82\%} \\ \bottomrule
\end{tabular}
\begin{tablenotes}%[para,flushleft]
        \item[a] Other lines except LR use PReLU as activation function.
      \end{tablenotes}
      \end{threeparttable}
\label{table:exptablePublic}
\end{table}
\vspace*{-0.4cm}

\subsection{Result from model comparison on Amazon Dataset and MovieLens Dataset}
Table \ref{table:exptablePublic} shows the results on Amazon dataset and MovieLens dataset.
%All the experiments are repeated 5 times with different seed, and the influence of random initialization on AUC is less than 0.0002.
All experiments are repeated 5 times and averaged results are reported. The influence of random initialization on AUC is less than 0.0002.
Obviously, all the deep networks beat LR model significantly, which indeed demonstrates the power of deep learning.   
PNN and DeepFM with specially designed structures preform better than Wide\&Deep. 
DIN performs best among all the competitors. 
Especially on Amazon Dataset with rich user behaviors, DIN stands out significantly.
We owe this to the design of local activation unit structure in DIN.
DIN pays attentions to the locally related user interests by soft-searching for parts of user behaviors that are relevant to candidate ad. With this mechanism, DIN obtains an adaptively varying representation of user interests, greatly improving the expressive ability of model compared with other deep networks.  
Besides, DIN with Dice brings further improvement over DIN, which verifies the effectiveness of the proposed data adaptive activation function Dice.

\begin{table}[]
\caption{Best AUCs of BaseModel with different regularizations on Alibaba Dataset corresponding to Fig.\ref{fig:adaptive_reg}. All the other lines calculate RelaImpr by comparing with first line.}
\small
\centering
\begin{threeparttable}
\begin{tabular}{p{5cm} p{0.6cm}<{\centering} p{0.8cm}<{\centering}}
\toprule
 Regularization    & AUC  &RelaImpr \\ \midrule
Without goods\_ids feature and Reg. & 0.5940 & 0.00\% \\
With goods\_ids feature without Reg. & 0.5959  & 2.02\% \\
With goods\_ids feature and Dropout Reg. & 0.5970  & 3.19\%  \\
With goods\_ids feature and Filter Reg.  & 0.5983& 4.57\%  \\
With goods\_ids feature and Difacto Reg.  & 0.5954 & 1.49\%  \\
\textbf{With goods\_ids feature and MBA. Reg.}  & \textbf{0.6031}  & \textbf{9.68\%}  \\ \bottomrule
\end{tabular}
\end{threeparttable}
\label{table:expreg}
\end{table}

\subsection{Performance of regularization}
\label{sec:reg}
As the dimension of features in both Amazon Dataset and MovieLens Dataset is not high (about 0.1 million), 
all the deep models including our proposed DIN do not meet grave problem of overfitting.
However, when it comes to the Alibaba dataset from the online advertising system which contains higher dimensional sparse features, overfitting turns to be a big challenge. 
For example, when training deep models with fine-grained features (e.g., features of $goods\_ids$ with dimension of 0.6 billion in Table \ref{table_feature_set}), serious overfitting occurs after the first epoch without any regularization, which causes the model performance to drop rapidly, as the dark green line shown in Fig.\ref{fig:adaptive_reg}.     
For this reason, we conduct careful experiments to check the performance of several commonly used regularizations.
%Here we compare several commonly used regularizations.
\begin{itemize}
\item \textbf{Dropout\cite{dropout}}. Randomly discard $50\%$ of feature ids in each sample.
\item \textbf{Filter}. Filter visited $goods\_id$ by occurrence frequency in samples and leave only the most frequent ones. In our setting, top $20$ million $goods\_ids$ are left.
\item \textbf{Regularization in DiFacto\cite{difacto}}. Parameters associated with frequent features are less over-regularized.
\item \textbf{MBA}. Our proposed \textbf{M}ini-\textbf{B}atch \textbf{A}ware regularization method (Eq.\ref{eq:batch}). Regularization parameter $\lambda$ for both DiFacto and MBA is searched and set to be $0.01$.
\end{itemize}

Fig.\ref{fig:adaptive_reg} and Table \ref{table:expreg} give the comparison results. 
Focusing on the detail of Fig.\ref{fig:adaptive_reg}, model trained with fine-grained $goods\_ids$ features brings large improvement on the test AUC performance in the first epoch, compared without it. 
However, overfitting occurs rapidly in the case of training without regularization (dark green line).   
Dropout prevents quick overfitting but causes slower convergence. 
Frequency filter relieves overfitting to a degree.
Regularization in DiFacto sets a greater penalty on $goods\_id$ with high frequency, which performs worse than frequency filter.
Our proposed mini-batch aware(MBA) regularization performs best compared with all the other methods, which prevents overfitting significantly. 

Besides, well trained models with $goods\_ids$ features show better AUC performance than without them. This is duo to the richer information that fine-grained features contained.
Considering this, although frequency filter performs slightly better than dropout, it throws away most of low frequent ids and may lose room for models to make better use of fine-grained features.

\subsection{Result from model comparison on Alibaba Dataset}
Table \ref{table:ali} shows the experimental results on Alibaba dataset with full feature sets as shown in Table \ref{table_feature_set}.
As expected, LR is proven to be much weaker than deep models. 
Making comparisons among deep models, we report several conclusions.  
First, under the same activation function and regularization, DIN itself has achieved superior performance compared with all the other deep networks including BaseModel, Wide\&Deep, PNN and DeepFM. DIN achieves 0.0059 absolute AUC gain and $6.08\%$ RelaImpr over BaseModel. It validates again the useful design of local activation unit structure.   
Second, ablation study based on DIN demonstrates the effectiveness of our proposed training techniques. Training DIN with mini-batch aware regularizer brings additional 0.0031 absolute AUC gain over dropout. Besides, DIN with Dice brings additional 0.0015 absolute AUC gain over PReLU. 
%In all, training DIN with the two proposed techniques together brings total 0.0054 AUC gain. 

Taken together, DIN with MBA regularization and Dice achieves total $11.65\%$ RelaImpr and 0.0113 absolute AUC gain over BaseModel. Even compared with competitor DeepFM which performs best on this dataset, DIN still achieves 0.009  absolute AUC gain. 
It is notable that in commercial advertising systems with hundreds of millions of traffics, 0.001 absolute AUC gain is significant and worthy of model deployment empirically. 
DIN shows great superiority to better understand and make use of the characteristics of user behavior data. 
Besides, the two proposed techniques further improve model performance and provide powerful help for training large scale industrial deep networks.   

\begin{table}[]
\caption{Model Comparison on Alibaba Dataset with full feature sets. All the lines calculate RelaImpr by comparing with BaseModel. 
DIN significantly outperforms all the other competitors. Besides, training DIN with our proposed mini-batch aware regularizer and Dice activation function brings further improvements.}
\small
\centering
\begin{threeparttable}
\begin{tabular}{p{4cm} p{1.5cm}<{\centering} p{1.5cm}<{\centering}}
\toprule
Model & AUC & RelaImpr \\ \midrule
LR                                  & 0.5738 &  - 23.92\%  \\    
BaseModel\tnote{a,b}               & 0.5970 &  0.00\% \\
Wide\&Deep\tnote{a,b}               & 0.5977 &  0.72\%  \\
PNN\tnote{a,b}                      & 0.5983 &  1.34\%  \\
DeepFM\tnote{a,b}                   & 0.5993 &  2.37\%  \\
\textbf{DIN Model\tnote{a,b}}       & \textbf{0.6029} & \textbf{6.08\%} \\
\textbf{DIN with MBA Reg.\tnote{a}} & \textbf{0.6060} &  \textbf{9.28}\%  \\
\textbf{DIN with Dice \tnote{b}}    & \textbf{0.6044} &  \textbf{7.63}\%  \\
\textbf{DIN with MBA Reg. and Dice} & \textbf{0.6083} &  \textbf{11.65\%}  \\ \bottomrule
\end{tabular}
\begin{tablenotes}%[para,flushleft]
        \item[a] These lines are trained with PReLU as the activation function. 
        \item[b] These lines are trained with dropout regularization. 
      \end{tablenotes}
      \end{threeparttable}
\label{table:ali}
\end{table}

\subsection{Result from online A/B testing}
Careful online A/B testing in the display advertising system in Alibaba was conducted from 2017-05 to 2017-06. 
During almost a month's testing,  DIN trained with the proposed regularizer and activation function contributes up to 10.0\% CTR and 3.8\% RPM(Revenue Per Mille) promotion\footnote{In our real advertising system, ads are ranked by $\textsl{CTR}^\alpha \cdot \textsl{bid-price}$ with $\alpha > 1.0$, which controls the balance of promotion of CTR and RPM.} compared with the introduced BaseModel, the last version of our online-serving model. This is a significant improvement and demonstrates the effectiveness of our proposed approaches. Now DIN has been deployed online and serves the main traffic.  

It is worth mentioning that online serving of industrial deep networks is not an easy job with hundreds of millions of users visiting our system everyday. Even worse, at traffic peak our system serves more than 1 million users per second. It is required to make realtime CTR predictions with high throughput and low latency. For example, in our real system we need to predict hundreds of ads for each visitor in less than 10 milliseconds.  
In our practice, several important techniques are deployed for accelerating online serving of industrial deep networks under the CPU-GPU architecture:  
i) \textsl{request batching} which merges adjacent requests from CPU to take advantage of GPU power,
ii) \textsl{GPU memory optimization} which improves the access pattern to reduce wasted transactions in GPU memory,
iii) \textsl{concurrent kernel computation} which allows execution of matrix computations to be processed with multiple CUDA kernels concurrently.           
In all, optimization of these techniques doubles the QPS (Query Per Second) capacity of a single machine practically. 
Online serving of DIN also benefits from this.

\subsection{Visualization of DIN}
\label{visual_din}
\begin{figure}[!h]
\centering
\includegraphics[height=2.5in, width=3.5in, keepaspectratio]{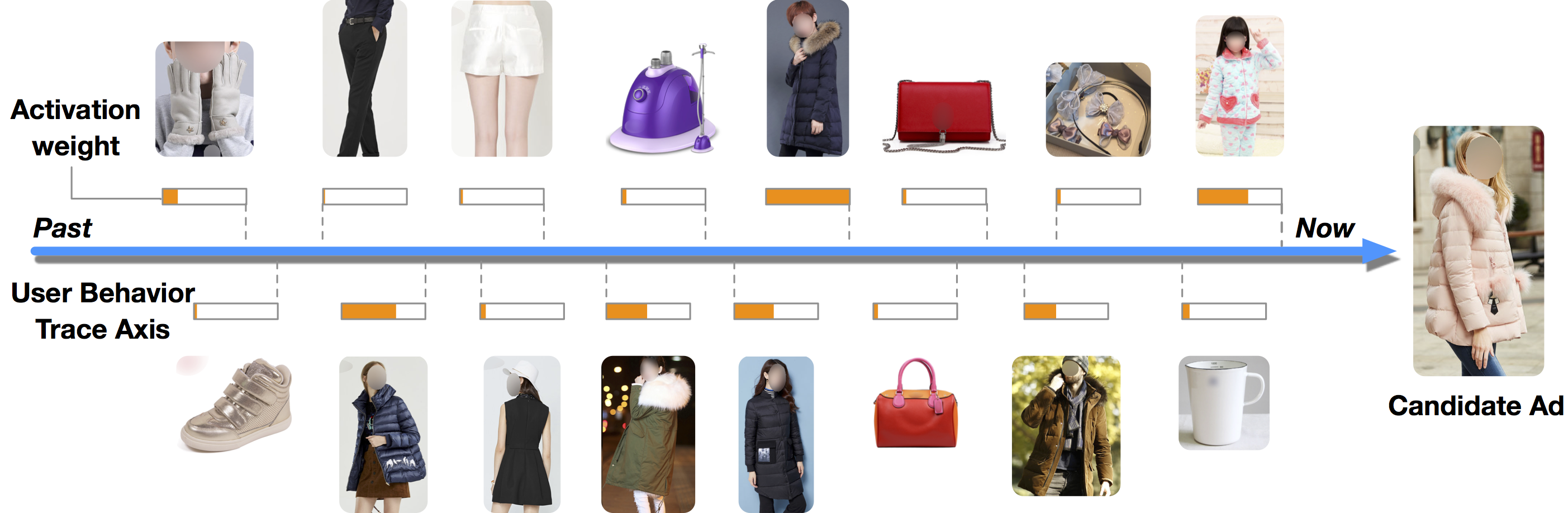}
\caption{Illustration of adaptive activation in DIN. Behaviors with high relevance to candidate ad get high activation weight. }
\label{fig:att_case}
%\end{figure*}
\end{figure}

Finally we conduct case study to reveal the inner structure of DIN on Alibaba dataset.
%DIN designs the activation unit to locally activate the related behaviors with respect to candidate ads. 
We first examine the effectiveness of local activation unit. 
Fig.\ref{fig:att_case} illustrates the activation intensity of user behaviors with respect to a candidate ad.
As expected, behaviors with high relevance to candidate ad are weighted high.

%\begin{figure*}[!h]

\begin{figure}[!t]
\centering
\includegraphics[height=2.5in, width=3.5in, keepaspectratio]{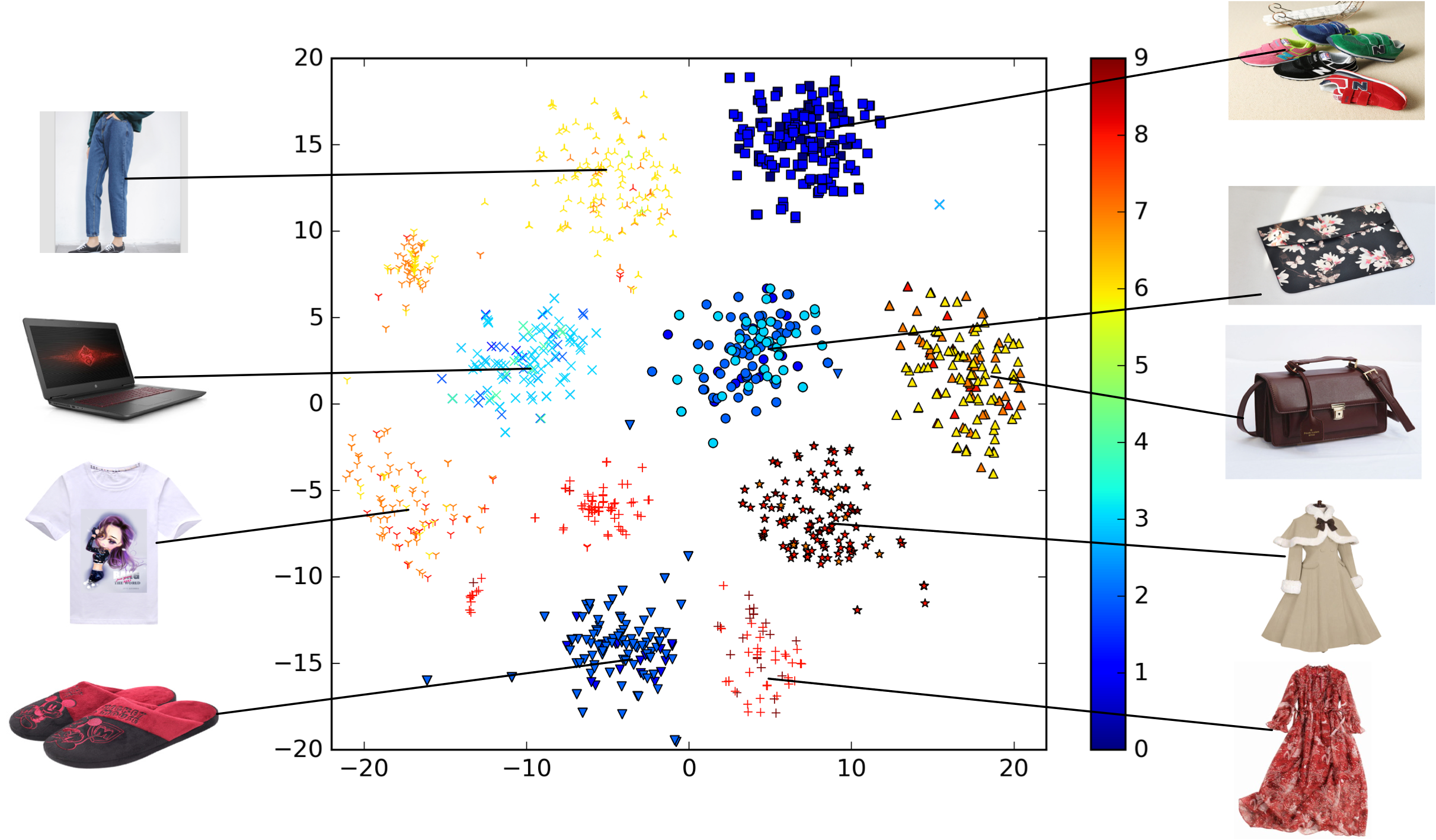}
\caption{Visualization of embeddings of goods in DIN. Shape of points represents category of goods. Color of points corresponds to CTR prediction value.}
\label{fig:TDdiagram}
\end{figure}

We then visualize the learned embedding vectors.
Taking the young mother mentioned before as example, we randomly select 9 categories (dress, sport shoes, bags, etc) and 100 goods of each category as the candidate ads for her.
Fig.\ref{fig:TDdiagram} shows the visualization of embedding vectors of goods with t-SNE\cite{tsne} learned by DIN, in which points with same shape correspond to the same category. 
We can see that goods with same category almost belong to one cluster, which shows the clustering property of DIN embeddings clearly.
Besides, we color the points that represent candidate ads by the prediction value. Fig.\ref{fig:TDdiagram} is also a heat map of this mother's interest density distribution for potential  candidates in embedding space. It shows DIN can form a multimodal interest density distribution in candidates' embedding space for a certain user to capture his/her diverse interests.

%Besides, we color the points that represent candidate ads by the prediction value. 
%Hot(red) points get higher CTR than cold(blue) ones.
%The red points appear in more than one clusters.
%Different categories of goods appearing the her historical behaviors are colored hot(red), which get higher CTR than those random candidates with cold(blue) color.  
%This demonstrates that DIN is able to capture diverse interests of a certain user, by forming a multimodal probability distribution of user interests in the embedding space.

%% file: ch_concln.tex
\section{Conclusions}
In this paper, we focus on the task of CTR prediction modeling in the scenario of display advertising in e-commerce industry with rich user behavior data.
The use of fixed-length representation in traditional deep CTR models is a bottleneck for capturing the diversity of user interests. 
To improve the expressive ability of model, a novel approach named DIN is designed to activate related user behaviors and obtain an adaptive representation vector for user interests which varies over different ads.    
%soft-searching for a part of user's behaviors to represent user's activated interests with a vector, which varies with candidate ads.\par
%With exploiting these characteristic sufficiently, we design a novel model named DIN.
%Besides, we propose a mini-batch aware regularization technique which can help reducing overfitting greatly in our scenario. What's more, we design a mini-batch aware activation function to accelerate convergence of our model. These approaches could be generalized to other industrial deep learning tasks.
Besides two novel techniques are introduced to help training industrial deep networks and further improve the performance of DIN. They can be easily generalized to other industrial deep learning tasks. 
DIN now has been deployed in the online display advertising system in Alibaba.  